\documentclass[10pt,logo,copyright]{giga-report}
\usepackage[authoryear,sort&compress,round]{natbib}

\usepackage[utf8]{inputenc} %
\usepackage[T1]{fontenc}    %

\usepackage{parskip}        %
\usepackage{url}            %
\usepackage{booktabs}       %
\usepackage{amsfonts}       %
\usepackage{nicefrac}       %
\usepackage{microtype}      %
\usepackage{xcolor}         %
\usepackage[dvipsnames]{xcolor} %
\usepackage{graphicx}
\usepackage{animate}        %
\usepackage{subcaption}
\usepackage{tabularx}
\usepackage{makecell}
\usepackage{adjustbox}
\usepackage{setspace}
\usepackage{todonotes}
\usepackage{colortbl}
\usepackage{wrapfig}

\newcolumntype{M}[1]{>{\centering\arraybackslash}m{#1}}
\usepackage{float}
\usepackage{tikz}
\usetikzlibrary{positioning,shapes,arrows}
\usepackage{amsmath,amsfonts,bm, bbm,leftindex}
\usepackage{multirow}
\usepackage{comment}
\usepackage{gensymb}
\usepackage{lipsum}
\usetikzlibrary{arrows.meta, positioning, fit}
\usepackage[para]{threeparttable}
\usepackage{tikz}
\usetikzlibrary{tikzmark}

\def\eqref#1{equation~\ref{#1}}

\def\1{\bm{1}}

\DeclareMathAlphabet{\mathsfit}{\encodingdefault}{\sfdefault}{m}{sl}
\SetMathAlphabet{\mathsfit}{bold}{\encodingdefault}{\sfdefault}{bx}{n}

\makeatletter
\let\save@mathaccent\mathaccent
\newcommand*\if@single[3]{%
  \setbox0\hbox{${\mathaccent"0362{#1}}^H$}%
  \setbox2\hbox{${\mathaccent"0362{\kern0pt#1}}^H$}%
  \ifdim\ht0=\ht2 #3\else #2\fi
  }
\newcommand*\rel@kern[1]{\kern#1\dimexpr\macc@kerna}
\newcommand*\widebar[1]{\@ifnextchar^{{\wide@bar{#1}{0}}}{\wide@bar{#1}{1}}}
\newcommand*\wide@bar[2]{\if@single{#1}{\wide@bar@{#1}{#2}{1}}{\wide@bar@{#1}{#2}{2}}}
\newcommand*\wide@bar@[3]{%
  \begingroup
  \def\mathaccent##1##2{%
    \let\mathaccent\save@mathaccent
    \if#32 \let\macc@nucleus\first@char \fi
    \setbox\z@\hbox{$\macc@style{\macc@nucleus}_{}$}%
    \setbox\tw@\hbox{$\macc@style{\macc@nucleus}{}_{}$}%
    \dimen@\wd\tw@
    \advance\dimen@-\wd\z@
    \divide\dimen@ 3
    \@tempdima\wd\tw@
    \advance\@tempdima-\scriptspace
    \divide\@tempdima 10
    \advance\dimen@-\@tempdima
    \ifdim\dimen@>\z@ \dimen@0pt\fi
    \rel@kern{0.6}\kern-\dimen@
    \if#31
      \overline{\rel@kern{-0.6}\kern\dimen@\macc@nucleus\rel@kern{0.4}\kern\dimen@}%
      \advance\dimen@0.4\dimexpr\macc@kerna
      \let\final@kern#2%
      \ifdim\dimen@<\z@ \let\final@kern1\fi
      \if\final@kern1 \kern-\dimen@\fi
    \else
      \overline{\rel@kern{-0.6}\kern\dimen@#1}%
    \fi
  }%
  \macc@depth\@ne
  \let\math@bgroup\@empty \let\math@egroup\macc@set@skewchar
  \mathsurround\z@ \frozen@everymath{\mathgroup\macc@group\relax}%
  \macc@set@skewchar\relax
  \let\mathaccentV\macc@nested@a
  \if#31
    \macc@nested@a\relax111{#1}%
  \else
    \def\gobble@till@marker##1\endmarker{}%
    \futurelet\first@char\gobble@till@marker#1\endmarker
    \ifcat\noexpand\first@char A\else
      \def\first@char{}%
    \fi
    \macc@nested@a\relax111{\first@char}%
  \fi
  \endgroup
}
\makeatother

\definecolor{darkred}{rgb}{0.7, 0.0, 0.0}

\usepackage{amsmath} 
\usepackage{marvosym}

\usepackage[table]{xcolor}

\usepackage{pifont}
\usepackage{graphicx}

\usepackage[nameinlink]{cleveref}
\crefname{equation}{Eq.}{Eqs.}
\crefname{figure}{Fig.}{Figs.}
\crefname{section}{Sec.}{Sec.}
\crefname{appendix}{App.}{App.}
\crefname{table}{Tab.}{Tabs.}
\crefname{algorithm}{Algo}{Algo}
\crefname{thm}{Thm}{Thm}
\Crefname{thm}{Thm}{Thm}
\crefname{prop}{Prop}{Prop}

\newcommand{\crefnames}[3]{%
  \@for\next:=#1\do{%
    \expandafter\crefname\expandafter{\next}{#2}{#3}%
  }%
}

\title{UniDriveDreamer: A Single-Stage Multimodal World Model for Autonomous Driving}

\author{
\centering{
Guosheng Zhao$^{1,2*}$ ~
Yaozeng Wang$^{1*}$ ~
Xiaofeng Wang$^{1*}$ ~
Zheng Zhu$^{1*}\textsuperscript{\Letter}$ ~
Tingdong Yu$^{1}$ ~
Guan Huang$^{1}$ ~
Yongchen Zai$^{3}$ ~
Ji Jiao$^{3}$ ~
Changliang Xue$^{3}$ ~
Xiaole Wang$^{3}$ ~
Zhen Yang$^{3}$ ~
Futang Zhu$^{3}$ ~
Xingang Wang$^{2}$\textsuperscript{\Letter}\\
\textbf{$^\text{1 }$GigaAI~~~~$^\text{2 }$CASIA~~~$^\text{3 }$BYD}}
}

\begin{document}
\maketitle

\begin{abstract}
World models have demonstrated significant promise for data synthesis in autonomous driving. However, existing methods predominantly concentrate on single-modality generation, typically focusing on either multi-camera video or LiDAR sequence synthesis. In this paper, we propose \textit{UniDriveDreamer}, a single-stage unified multimodal world model for autonomous driving, which directly generates multimodal future observations without relying on intermediate representations or cascaded modules. Our framework introduces a LiDAR-specific variational autoencoder (VAE) designed to encode input LiDAR sequences, alongside a video VAE for multi-camera images. To ensure cross-modal compatibility and training stability, we propose Unified Latent Anchoring (ULA), which explicitly aligns the latent distributions of the two modalities. The aligned features are fused and processed by a diffusion transformer that jointly models their geometric correspondence and temporal evolution. Additionally, structured scene layout information is projected per modality as a conditioning signal to guide the synthesis. Extensive experiments demonstrate that \textit{UniDriveDreamer} outperforms previous state-of-the-art methods in both video and LiDAR generation, while also yielding measurable improvements in downstream driving tasks.
\end{abstract}

\abscontent
\section{Introduction}
\label{sec:intro}

World models \citep{drivedreamer, gaia1, drivewm, vista, drivedreamer4d, recondreamer, recondreamer++, cosmos, cosmosdrivedream, cosmos2.5, gigaworld, worldmodel, recondreamerrl} have made significant progress in autonomous driving, and their powerful data synthesis capabilities offer a promising solution for reducing the costs of data acquisition and annotation. Recent studies \citep{magicdrive, gaia2, magicdrivev2, panacea, instadrive} have established a notable paradigm that leverages structured scene layouts, such as 3D bounding boxes and HDMaps, as conditional inputs for driving data generation. The synthesized data can be further utilized to enhance the performance of downstream tasks \citep{bevformer, bevfusion}.

Nevertheless, existing methods primarily focus on single-modality generation, such as camera-only RGB synthesis \citep{drivedreamer, drivedreamer2, drivewm, magicdrive, magicdrivev2, panacea, vista, drivingdiff, bevcontrol} or LiDAR sequence generation \citep{lidargen, lidardm, rangeldm, lidarvae}. Although these approaches are able to produce highly realistic data that benefit downstream perception tasks, the inherent single-modality nature limits their applicability in scenarios demanding diverse multi-sensor observations. Recently, a few efforts have explored multimodal generation for autonomous driving, demonstrating the concrete potential of multimodal data synthesis to improve downstream task performance. However, existing methods \citep{uniscene, genesis} typically decouple the generation of different modalities. In such frameworks, the synthesis of one modality (e.g., LiDAR) is explicitly conditioned on the output or latent representation of another (e.g., RGB). This decoupled architecture lacks explicit, bidirectional interaction between the RGB and LiDAR modalities during the synthesis process, which consequently undermines cross-modal consistency. The resulting unidirectional information flow impedes the effective utilization of complementary sensor data, consequently degrading the overall quality and coherence of the generated multimodal outputs.

To address this challenge, we introduce \textit{UniDriveDreamer}, a single-stage multimodal world model designed for autonomous driving, which enables deep fusion and bidirectional interaction across different modalities, thereby substantially enhancing the spatiotemporal and cross-modal consistency of synthesized outputs. Specifically, a LiDAR-specific variational autoencoder (VAE) is designed to encode input range images into a latent space, while multi-view images are processed by a video VAE \citep{wan, vae}. Furthermore, to stabilize training and ensure compatibility between modalities, we propose Unified Latent Anchoring (ULA), which aligns the distribution of LiDAR latents with the prior of a pretrained RGB VAE via affine transformations derived from empirical statistics. The aligned latent features are then fused and processed by a diffusion transformer, which jointly models both intra- and cross-modal spatiotemporal relationships. In addition, we incorporate structured scene layout information to guide the generation process by projecting it into each modality as a conditioning signal. Extensive experimental results demonstrate that \textit{UniDriveDreamer} surpasses previous state-of-the-art methods (SOTA) in multimodal data synthesis quality. Specifically, for video generation, \textit{UniDriveDreamer} achieves an FID of 2.81 and an FVD of 11.44. For LiDAR synthesis, it attains an MMD of 0.27 and a JSD of 0.039, corresponding to 82.3\% and 45.8\% relative improvements over UniScene \citep{uniscene}. In addition, downstream task evaluations further confirm the practical effectiveness of the generated data, showing relative improvements of +1.2\% in mAP and +0.7\% in NDS.

Overall, our contributions can be summarized as follows:
\begin{itemize}
    \item We propose \textit{UniDriveDreamer}, a single-stage multimodal world model for autonomous driving, which directly generates temporally consistent and geometrically coherent future observations, including multi-camera videos and LiDAR sweeps, without relying on intermediate representations or cascaded modules.
    \item We propose a LiDAR-specific VAE, capable of accurately reconstructing LiDAR sequences. Furthermore, to ensure cross-modal compatibility, Unified Latent Anchoring (ULA) is proposed to align their distributions via affine transformations based on empirical statistics, thereby stabilizing training and ensuring cross-modal compatibility.
    \item Experimental results demonstrate that \textit{UniDriveDreamer} not only surpasses previous SOTA methods in multimodal synthesis quality but also delivers significant improvements in downstream perception task performance.
\end{itemize}
\section{Related Work}
\subsection{Unimodal Driving Scene Generation}
The rapid advancement of diffusion models \citep{sd, svd, controlnet, worlddreamer, genie3, hunyuan, wan, cosmos, cosmos2.5, drivegen3d, zhu2024sora} has catalyzed a surge of research in driving video generation \citep{drivedreamer, drivedreamer2, instadrive, magicdrive, magicdrivev2, vista, panacea, gaia1, gaia2, cosmosdrivedream, cosmostransfer, drivingdiff, bevgen}. These methods commonly employ structured scene layouts as conditional inputs to synthesize realistic driving videos, substantially lowering the cost of data acquisition and annotation while demonstrating effectiveness in enhancing downstream perception performance. Furthermore, several approaches \citep{drivewm, delphi, genad} have investigated the application of world models in end-to-end autonomous driving. In parallel, as LiDAR serves as a crucial sensing modality in driving scenarios, recent methods \citep{lidardm, rangeldm, lidargen, lidarvae, cosmosdrivedream} have begun to introduce generative techniques for LiDAR data synthesis. These methods typically represent LiDAR data as range images by transforming the 3D point clouds to 2D pixel space, thereby aligning its format with that of RGB images. This representation facilitates the application of well-established image and video generation techniques, leading to significant advancements in the field. Despite these advances, existing approaches remain predominantly focus on single-modality generation, thereby limiting the realization of sensor-complete autonomous driving simulation environments. To bridge this gap, we propose \textit{UniDriveDreamer}, a unified world model designed to address multimodal data synthesis in driving scenarios, capable of jointly generating multi-view camera videos and LiDAR sequences.

\subsection{Multimodal Driving Scene Generation}
To meet the demand for multimodal data in downstream tasks, recent works \citep{omnigen, genesis, uniscene, bevworld, unifuture} have begun to explore multimodal data generation for driving scenarios. For instance, UniScene \citep{uniscene} proposes a two-stage pipeline that first generates an occupancy representation, then synthesizes camera and LiDAR data conditioned on it. Following a similar cascaded philosophy, Genesis \citep{genesis} employs a dual-branch architecture, where the explicit occupancy is replaced by a latent video representation extracted from the RGB branch, and LiDAR generation is conditioned on these features sequentially. Despite these advances, such cascaded or sequential paradigms lack explicit bidirectional interaction between RGB and LiDAR during synthesis, which constrains the overall quality and consistency of the generated multimodal outputs. In contrast, our \textit{UniDriveDreamer} adopts a single-stage architecture that enables bidirectional interaction between RGB and LiDAR modalities throughout the generation process, facilitating deeper fusion and more coherent multimodal synthesis.
\begin{figure}[t]
\centering
\captionsetup{type=figure, justification=justified, singlelinecheck=false}
\includegraphics[width=\textwidth]{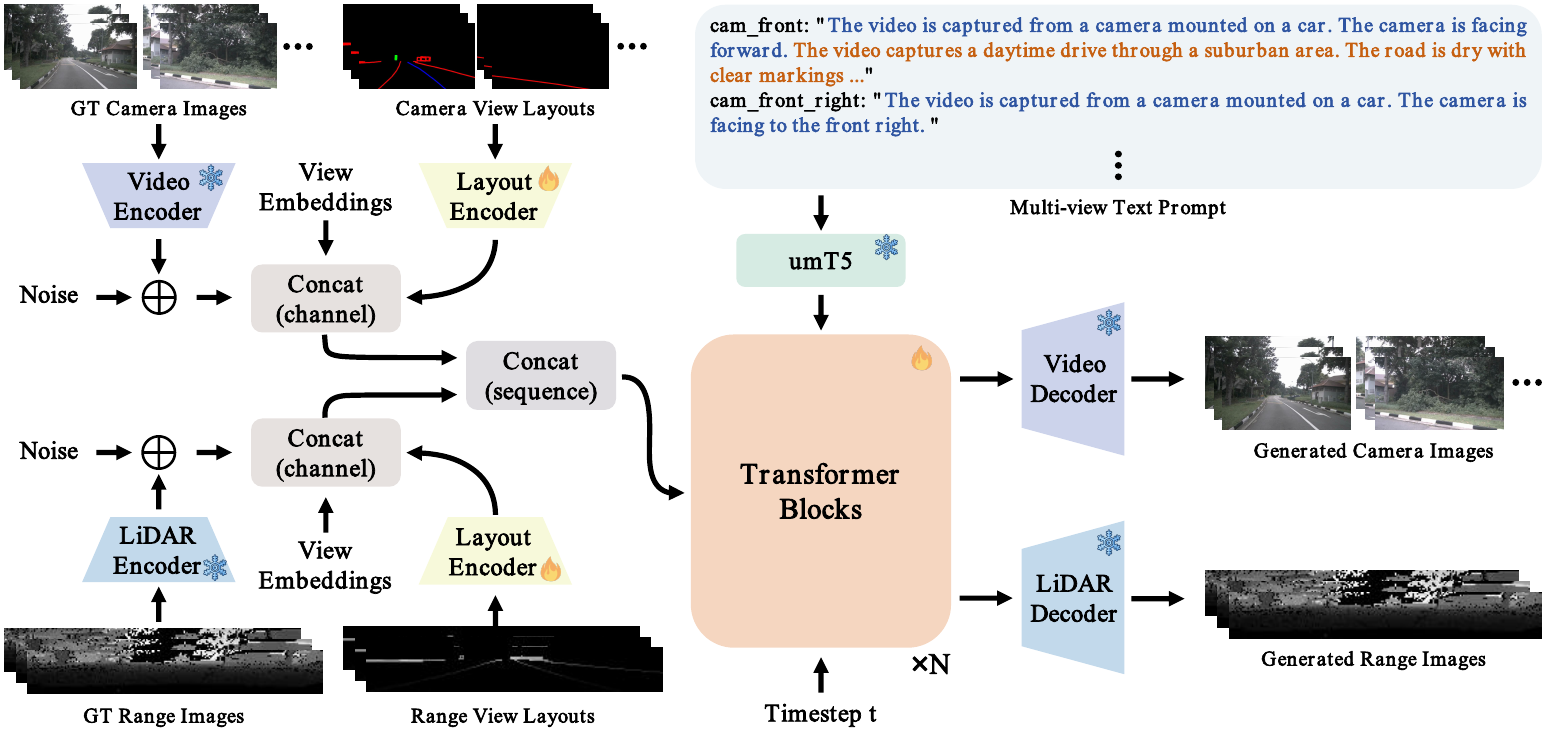}
\caption{The overall framework of \textit{UniDriveDreamer}. Our \textit{UniDriveDreamer} consists of four core components: (1) two modality-specific VAEs that encode multi-view camera images and LiDAR range maps into a shared latent space; (2) a layout encoder that projects structured scene layout information into corresponding latent representations; (3) a text encoder that encode multi-view text prompts into corresponding prompt embeddings; and (4) a diffusion transformer that jointly models spatiotemporal coherence within each modality and cross-modal consistency across modalities.}
\label{fig_framework}
\end{figure}

\section{UniDriveDreamer}
The overall framework is illustrated in Fig.~{\ref{fig_framework}}. Multi-view images and LiDAR range maps are first encoded by their dedicated VAEs. The resulting latent features are then concatenated in the channel dimension with the corresponding modality-specific layout latent features and view embeddings. Next, the two sets of multimodal latent tokens are concatenated along the sequence dimension and fed directly into a diffusion transformer, which jointly models both intra-modal spatiotemporal consistency and cross-modal alignment in the latent space. Finally, the output latent representations are decoded back into their respective modalities, RGB video frames and LiDAR range sequences, through the corresponding decoder networks.

\subsection{LiDAR VAE}
In this section, we detail the training and inference process of the proposed LiDAR‑specific VAE. Notably, to facilitate joint generation of LiDAR sequences and camera videos, we adopt range images as the LiDAR representation, thereby aligning its format more closely with RGB imagery.

\noindent \textbf{Training.} Our LiDAR VAE preserves the same architectural backbone as the RGB VAE proposed in \citep{wan}, differing only in the number of input and output channels, which are set to 1 instead of 3 to accommodate the single-channel nature of range images. Following \citep{cosmosdrivedream}, we further mitigate the inherent sparsity of range images by repeating each row of the range map four times, yielding an input sequence $v^{\text{L}} \in \mathbb{R}^{(1+T)\times H^{\text{L}}\times W^{\text{L}} \times 1 }$. The encoder of the LiDAR VAE maps the sequence to a latent representation $z^{\text{L}}=\xi(v^\text{L}) \in \mathbb{R}^{(1+\frac{T}{4})\times \frac{H^{\text{L}}}{8} \times \frac{W^{\text{L}}}{8}\times C}$, which is subsequently decoded back into the reconstructed LiDAR video $\hat v^{\text{L}} = \mathcal{G}(z^\text{L})\in \mathbb{R}^{(1+T)\times H^{\text{L}}\times W^{\text{L}} \times 1}$. The model is trained using the following composite loss function:
\begin{equation}
    \mathcal{L}_{\textbf{vae}} = \lambda_{1}\Vert v^\text{L} - \hat v^{\text{L}}\Vert_1 + \lambda_{2}D_{\text{KL}}(q(z^{\text{L}}|v^{\text{L}})\Vert p(z^{\text{L}})) + \lambda_{3} \mathcal{L}_{\text{LPIPS}}(v^{\text{L}}, \hat v^{\text{L}}),
\end{equation}
where $D_{\text{KL}}(\cdot)$ is the Kullback–Leibler divergence between the approximate posterior $q(z^{\text{L}}|v^{\text{L}})$ and the standard Gaussian prior $p(z^{\text{L}})\sim\mathcal{N}(z^{\text{L}};\textbf{0},\textbf{I})$, and $\mathcal{L}_{\text{LPIPS}}$ is the Learned Perceptual Image Patch Similarity (LPIPS) loss \citep{lpips}. Since LPIPS is originally defined for RGB images, we replicate the single-channel range images across three channels when computing this perceptual loss term.

\noindent \textbf{Inference.} Once trained, the LiDAR VAE is integrated into the multimodal world model for joint training. To stabilize training and ensure cross-modal compatibility, we introduce Unified Latent Anchoring (ULA), which aligns the latent distribution of LiDAR features with that of the camera modality. Let $\mu^{\text{C}}, \sigma^{\text{C}}$ be the normalization parameters of the pretrained RGB VAE, $\mu_1^{\text{C}}, \sigma_1^{\text{C}}$ denote the statistics computed from the current driving dataset using the RGB VAE, and $\mu_1^{\text{L}}, \sigma_1^{\text{L}}$ refer the corresponding statistics from the LiDAR VAE. The calibrated normalization parameters $\mu^{\text{L}}, \sigma^{\text{L}}$ for the LiDAR latent features are then obtained as:
\begin{align}
    &\quad \frac{\frac{z^{\text{L}}-\mu_1^{\text{L}}}{\sigma_1^{\text{L}}}\cdot\sigma_1^{\text{C}}+\mu_1^{\text{C}}-\mu^{\text{C}}}{\sigma^{\text{C}}} \notag\\
    \Rightarrow &\quad \frac{{z^{\text{L}}}-\left(\mu_1^{\text{L}}-\mu_1^{\text{C}}\frac{\sigma_1^\text{L}}{\sigma_1^{\text{C}}}+\mu^{\text{C}}\frac{\sigma_1^\text{L}}{\sigma_1^{\text{C}}}\right)}{\frac{\sigma_1^{\text{L}}\sigma^{\text{C}}}{\sigma_1^{\text{C}}}} \notag\\
    \Rightarrow &\quad \mu^{\text{L}} = \mu_1^{\text{L}}-\mu_1^{\text{C}}\frac{\sigma_1^\text{L}}{\sigma_1^{\text{C}}}+\mu^{\text{C}}\frac{\sigma_1^\text{L}}{\sigma_1^{\text{C}}},\quad \sigma^{\text{L}} = \frac{\sigma_1^{\text{L}}\sigma^{\text{C}}}{\sigma_1^{\text{C}}}.
\end{align}

\subsection{Single-Stage Multimodal World Model}
In this section, we present the detailed design of our single-stage multimodal world model, \textit{UniDriveDreamer}. As illustrated in Fig.~\ref{fig_framework}, \textit{UniDriveDreamer} comprises four key components: two modality-specific VAEs, a layout encoder, a text encoder, and a diffusion transformer. Given an input sequence of multi-camera videos $v^{\text{C}}\in \mathbb{R}^{V\times (1+T)\times H^{\text{C}} \times W^{\text{C}} \times 3}$ and LiDAR sweeps $v^{\text{L}}\in \mathbb{R}^{(1+T)\times H^{\text{L}}\times W^{\text{L}} \times 1}$, each modality is first encoded into the latent space $z^{\text{C}}\in \mathbb{R}^{V\times (1+\frac{T}{4})\times \frac{H^{\text{C}}}{8} \times \frac{W^{\text{C}}}{8} \times C}$, $z^{\text{L}}\in \mathbb{R}^{(1+\frac{T}{4})\times \frac{H^{\text{L}}}{8} \times \frac{W^{\text{L}}}{8} \times C}$ by its corresponding VAE. Multi‑view textual prompts are encoded using a pretrained text encoder \cite{umt5}. The layout encoder and the diffusion transformer are described in detail below.

\noindent \textbf{Layout Encoder.} The Layout Encoder consists of a spatial downsampling block followed by two spatiotemporal downsampling blocks \citep{wan}, aligning the spatiotemporal compression rate of layout latents with that of the multimodal data. Specifically, we first project structured 3D scene information, including 3D bounding boxes and HDMap, into both camera views and LiDAR range views. In camera views, different object categories and lane markings are distinguished by distinct colors, while in range views, the projection yields single-channel distance values. To enable the same layout encoder to process both modalities, we replicate the single-channel range-view layout map three times along the channel dimension before feeding it into the encoder. The encoder then outputs multi-camera view condition latents $c_{\text{layout}}^{\text{C}}\in \mathbb{R}^{V\times (1+\frac{T}{4})\times \frac{H^{\text{C}}}{8} \times \frac{W^{\text{C}}}{8} \times C}$ and range-view condition latents $c_{\text{layout}}^{\text{L}}\in \mathbb{R}^{(1+\frac{T}{4})\times \frac{H^{\text{L}}}{8} \times \frac{W^{\text{L}}}{8} \times C}$, which are then used to guide the respective modality synthesis.

\begin{figure}[!t]
\centering
\captionsetup{type=figure, justification=justified, singlelinecheck=false}
\includegraphics[width=\textwidth]{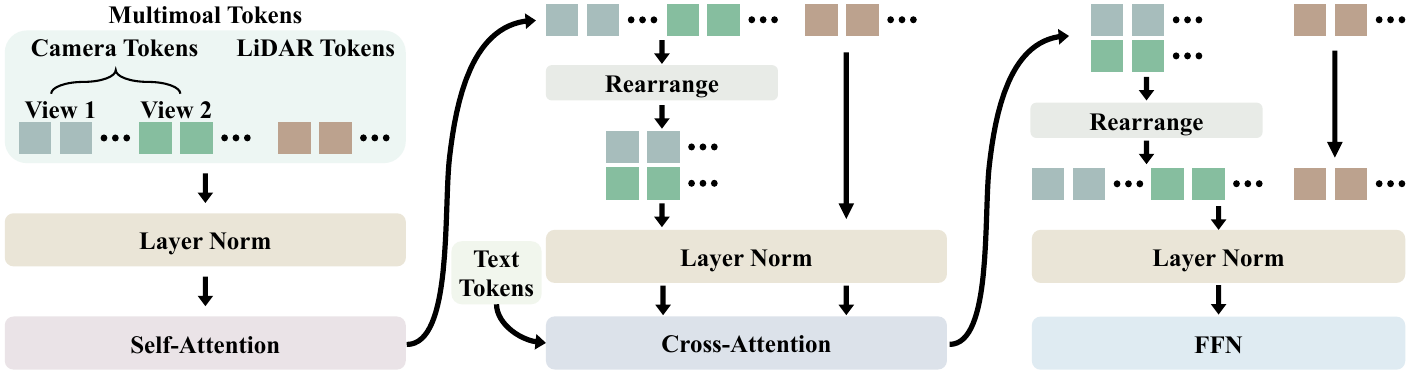}
\caption{Transformer block of \textit{UniDriveDreamer}.}
\label{fig_tb}
\end{figure}

\noindent \textbf{Diffusion Transformer.} The diffusion transformer architecture comprises three main components: (1) two modality-specific patchifying modules, (2) a shared stack of transformer blocks, and (3) two modality-specific unpatchifying modules. In the patchifying modules, a 3D convolution with a kernel size of (1, 2, 2) followed by a flattening operation is applied to convert the camera-views and range-view inputs into camera tokens $T^\text{C}$ and LiDAR tokens $T^\text{L}$ of shapes $\left(B, (V\ L^{\text{C}}), D\right)$ and $(B, L^{\text{L}}, D)$, respectively. The camera-view input is constructed by concatenating along the channel dimension the noisy latents $z^{\text{C}}_{\text{t}}$, view-specific embeddings $emb^{\text{C}}$, layout-conditioning latents $c_{\text{layout}}^{\text{C}}$, and first frame conditioning latents $c^{\text{C}}_{\text{frame}}$; the range-view input is formed analogously. These two sets of modality tokens are then concatenated along the sequence dimension and fed into the shared transformer block stack. As depicted in Fig.~\ref{fig_tb}, the concatenated multimodal tokens first pass through a self-attention module, which jointly models intra-modal spatiotemporal dependencies and enforces cross-modal consistency. The tokens are then separated by modality. The camera tokens are reshaped to $\left((B\ V), L^{\text{C}}, D\right)$ and processed together with the LiDAR tokens through a cross‑attention layer that conditions the generation on textual prompts. Subsequently, the camera tokens are reshaped back to their original layout and concatenated with the LiDAR tokens before being passed through a feed-forward network. After passing through the stack of transformer blocks, the modality-specific tokens are individually transformed by their corresponding unpatchifying modules to recover the original latent shape of each modality.

\noindent \textbf{Optimization.} \textit{UniDriveDreamer} is optimized by using the flow matching framework \citep{flowmatch, rectifiedflow}. Given a timestep $t\in [0,1]$ sampled from a logit-normal distribution, a random camera noise $z_0^{\text{C}}\in \mathbb{R}^{(1+\frac{T}{4})\times \frac{H^{\text{C}}}{8} \times \frac{W^{\text{C}}}{8} \times C}$, and a random range-view noise $z_0^{\text{L}} \in \mathbb{R}^{(1+\frac{T}{4})\times \frac{H^{\text{L}}}{8} \times \frac{W^{\text{L}}}{8} \times C}$, the noisy latent $z_t$ is obtained  as:
\begin{equation}
    z_t = t\begin{bmatrix}z_1^{\text{C}}\\z_1^{\text{L}}\end{bmatrix}
    +(1-t)\begin{bmatrix}z_0^{\text{C}}\\z_0^{\text{L}}\end{bmatrix},
\end{equation}
where $z_1^{\text{C}}$ and $z_1^{\text{L}}$ denote the clean camera and LiDAR latents, respectively. The corresponding ground-truth velocity $\nu_t$ is computed as:
\begin{equation}
    \nu_t=\frac{dz_t}{dt}=\begin{bmatrix}z_1^{\text{C}}-z_0^{\text{C}}\\z_1^{\text{L}}-z_0^{\text{L}}\end{bmatrix}.
\end{equation}
The model is then optimized to predict this velocity via the following objective:
\begin{equation}
    \mathcal{L} = \mathbb{E}_{z_1,z_0,c,t}\Vert u(z_t,c,t;\theta)-\nu_t\Vert^2,
\end{equation}
where the conditioning signal $c$ comprises layout conditions, multi-view text prompt embeddings, and first frame latents. The parameter set $\theta$ includes the weights of the diffusion transformer and the layout encoder.
\begin{table}[ht]
\caption{Comparison of the generation quality on nuScenes validation set.}
\centering
\begin{tabular}{@{}m{1.8cm}m{5.5cm} 
    >{\centering\arraybackslash}p{1.6cm} 
    >{\centering\arraybackslash}p{1.6cm} 
    >{\centering\arraybackslash}p{1.6cm} 
    >{\centering\arraybackslash}p{1.6cm}@{}}
\toprule
\multirow{2}{*}{Modality} & \multirow{2}{*}{Method} & \multicolumn{2}{c}{Camera Generation} & \multicolumn{2}{c}{LiDAR Generation} \\ 
\cmidrule(lr){3-4} \cmidrule(lr){5-6} 
& & FID $\downarrow$ & FVD $\downarrow$ & MMD $\downarrow$ & JSD $\downarrow$ \\ 
\midrule
\multirow{6}{*}{Camera} 
& DriveDreamer \citep{drivedreamer} & 14.90 & 340.80 & - & - \\
& DriveDreamer-2 \citep{drivedreamer2} & 11.20 & 55.70 & - & - \\
& MagicDrive \citep{magicdrive} & 16.20 & - & - & - \\
& MagicDrive-V2 \citep{magicdrivev2} & 20.91 & 94.84 & - & - \\
& Panacea \citep{panacea} & 16.96 & 139.0 & - & - \\
& Drive-WM \citep{drivewm} & 15.80 & 122.70 & - & - \\
\midrule
\multirow{3}{*}{LiDAR} 
& LiDARVAE \citep{lidarvae} & - & - & 11.0 & - \\
& LiDARGen \citep{lidargen} & - & - & 19.0 & 0.160 \\
& RangeLDM \citep{rangeldm} & - & - & 1.90 & 0.054 \\
\midrule
\multirow{4}{*}{Multimodal} 
& UniScene \citep{uniscene} & 6.12 & 70.52 & 1.53 & 0.072 \\
& OmniGen \citep{omnigen} & 21.01 & - & 2.94 & 0.105 \\
& Genesis \citep{genesis} & 4.24 & 16.95 & - & - \\
& \textit{UniDriveDreamer} & \textbf{2.81} & \textbf{11.44} & \textbf{0.27} & \textbf{0.039} \\
\bottomrule
\end{tabular}
\label{tab:gen_cmp}
\end{table}
\section{Experiment}
In this section, we present our experimental setup, including the datasets, implementation details and evaluation metrics. Subsequently, both quantitative and qualitative results are provided to demonstrate the superior performance of the proposed \textit{UniDriveDreamer}. In addition, ablation studies are conducted to validate key design choices.



\subsection{Experiment Setup}
\noindent \textbf{Datasets.} The training dataset is derived from the nuScenes dataset \citep{nuscenes}, consisting of 700 training sequences and 150 validation sequences. Each sequence encompasses approximately 20 seconds of recorded driving data, captured by six surround-view cameras and a roof-mounted LiDAR. Following \citep{drivedreamer, asap, drivedreamer2}, we preprocess the nuScenes dataset to calculate 12Hz annotations.

\noindent \textbf{Implementation Details.} In our experiments, camera videos are processed at a resolution of $432\times 768$ and LiDAR range maps at $32\times 1024$, with a temporal length of 17 frames. \textit{UniDriveDreamer} is initialized with the pretrained weights of Wan-2.1 T2V-1.3B \citep{wan}. The model is optimized with a learning rate of $2\times 10^{-5}$. For the LiDAR VAE, we load the pretrained VAE weights from Wan-2.1 \citep{wan} and set the learning rate to $5\times 10^{-5}$. The reconstruction loss, KL divergence loss, and LPIPS loss are weighted with coefficients $\lambda_1=1$, $\lambda_2=1$, and $\lambda_3=0.3$ respectively. Furthermore, following \citep{wan}, we adopt the umT5 model \citep{umt5} as the text encoder. 

\noindent \textbf{Evaluation Metrics.} 
The evaluation in this work covers three key aspects: generation quality, LiDAR reconstruction fidelity, and downstream task performance. To assess the quality of synthesized camera videos, we employ Fréchet Inception Distance (FID) \citep{fid} and Fréchet Video Distance (FVD) \citep{fvd}. For LiDAR sequences, we follow the evaluation protocol of \citep{lidardm} and adopt Maximum Mean Discrepancy (MMD) and Jensen–Shannon Divergence (JSD), where the MMD results are reported scaled by $10^4$. The geometric accuracy of LiDAR reconstruction is measured using Chamfer Distance and F‑Score as adopted in \citep{omnigen}. Finally, we employ BEVFusion \citep{bevfusion} to assess the utility of the generated data for downstream perception and report the nuScenes Detection Score (NDS) and mean Average Precision (mAP) for 3D object detection.

\begin{figure}[ht]
\centering
\captionsetup{type=figure, justification=justified, singlelinecheck=false}
\includegraphics[width=\textwidth]{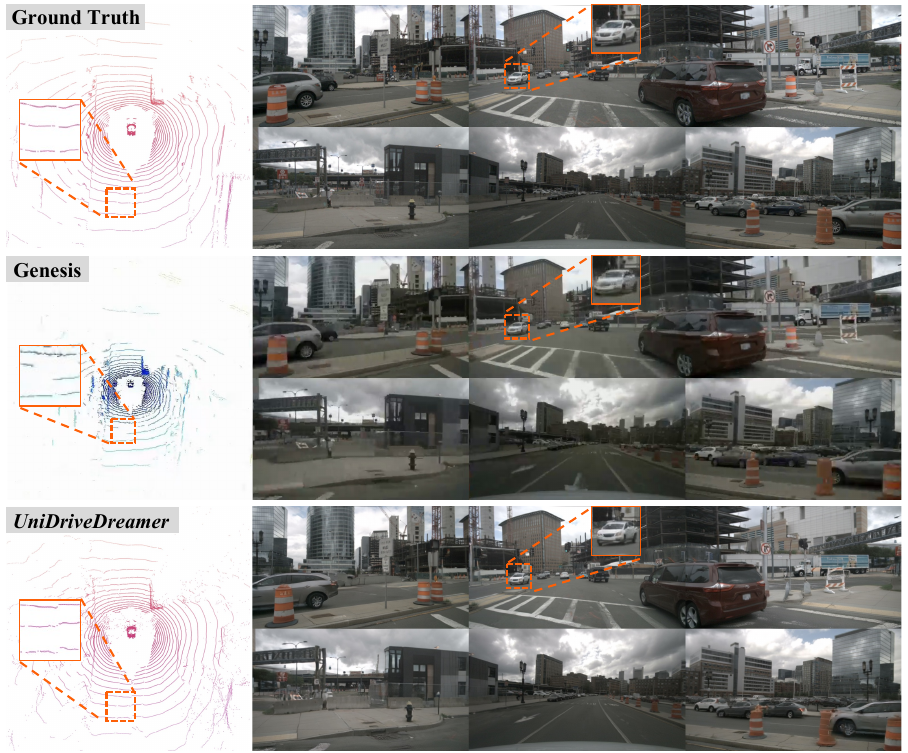}
\caption{Qualitative comparison of generated outputs with Genesis \citep{genesis}. The top row shows the ground truth. The middle row presents results from Genesis. The bottom row displays outputs from our \textit{UniDriveDreamer}.}
\label{fig:gen_cmp}
\end{figure}

\subsection{Main Results}
In this section, we present the experimental results in three aspects: (1) quality of multimodal data generation, (2) LiDAR reconstruction fidelity, and (3) downstream perception performance gains.

\begin{figure}[!h]
\centering
\captionsetup{type=figure, justification=justified, singlelinecheck=false}
\includegraphics[width=\textwidth]{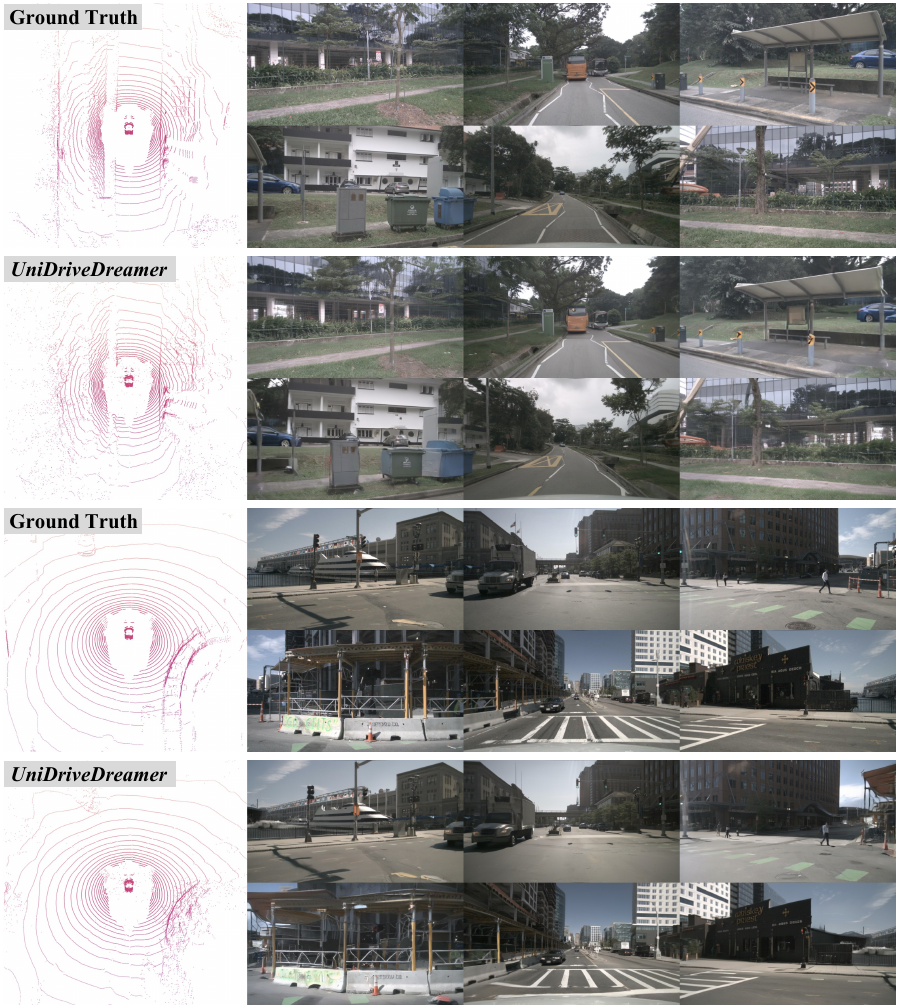}
\caption{Qualitative visualization of multimodal outputs generated by \textit{UniDriveDreamer}. In each example set, the upper row presents the ground truth, while the lower row shows the corresponding synthesized results generated by \textit{UniDriveDreamer}.}
\label{fig:quali_results_1}
\end{figure}

\begin{figure}[!h]
\centering
\captionsetup{type=figure, justification=justified, singlelinecheck=false}
\includegraphics[width=\textwidth]{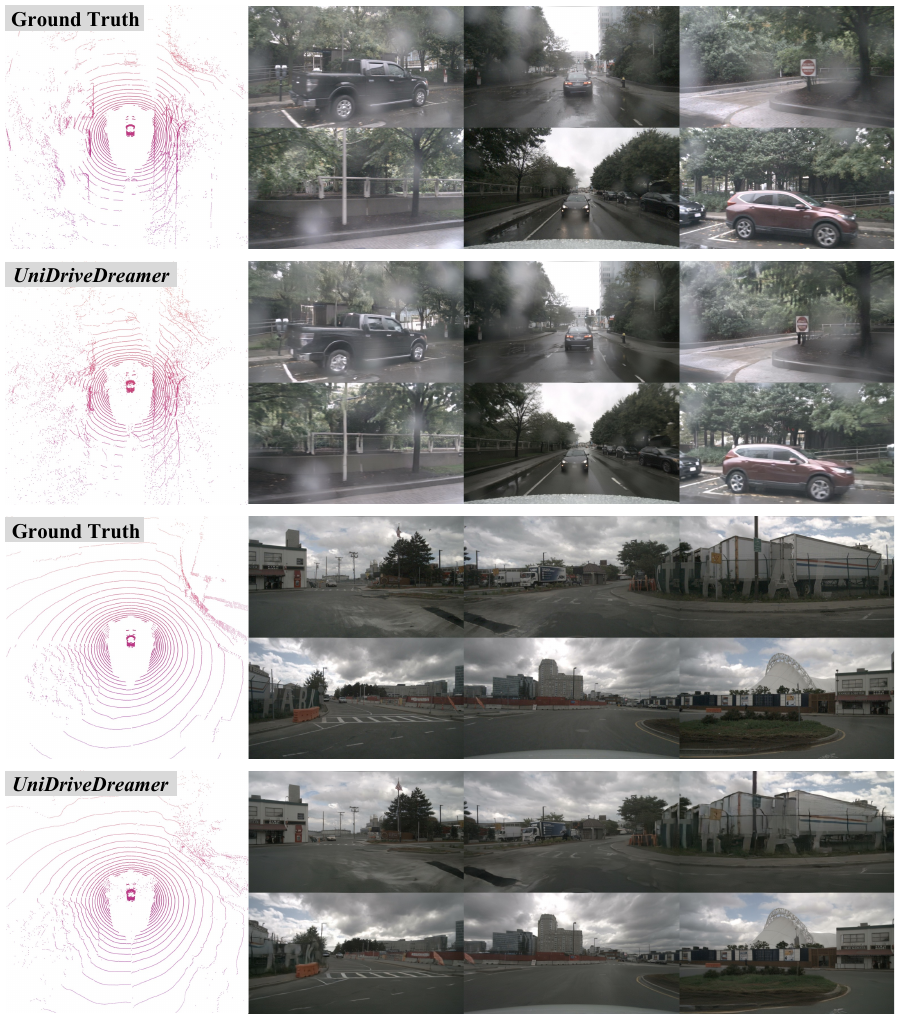}
\caption{Qualitative visualization of multimodal outputs generated by \textit{UniDriveDreamer}. In each example set, the upper row presents the ground truth, while the lower row shows the corresponding synthesized results generated by \textit{UniDriveDreamer}.}
\label{fig:quali_results_2}
\end{figure}

\noindent \textbf{Multimodal Generation Results.} 
As shown in Tab. \ref{tab:gen_cmp}, we compare \textit{UniDriveDreamer} against SOTA single-modal and multimodal methods \citep{drivedreamer, drivedreamer2, magicdrive, magicdrivev2, panacea, drivewm, lidarvae, lidargen, rangeldm, uniscene, omnigen, genesis} on the nuScenes validation set \citep{nuscenes} in terms of generation quality. Quantitative results demonstrate that \textit{UniDriveDreamer} achieves superior performance in both camera video and LiDAR sequence synthesis. Specifically, for video generation, \textit{UniDriveDreamer} attains an FID of 2.81 and an FVD of 11.44. For LiDAR synthesis, it yields an MMD of 0.27 and a JSD of 0.039, corresponding to improvements of 82.3\% in MMD and 45.8\% in JSD over UniScene \citep{uniscene}. 
Qualitative comparisons in Fig. \ref{fig:gen_cmp} demonstrate the superior generation quality of \textit{UniDriveDreamer}. For LiDAR synthesis, the results of \textit{UniDriveDreamer} exhibit more complete ground coverage and higher geometric fidelity than those of Genesis \citep{genesis}, showing closer alignment with the ground truth. For camera video generation, our method produces visually sharper and more realistic outputs with enhanced detail preservation. Additional generated examples are presented in Fig. \ref{fig:quali_results_1} and Fig. \ref{fig:quali_results_2}. These visual results further demonstrate the high controllability of our model under varied scene layouts, its strong spatiotemporal consistency across extended sequences, and its robust cross‑modal alignment between camera views and LiDAR sweeps.

\begin{figure}[!t]
\centering
\captionsetup{type=figure, justification=justified, singlelinecheck=false}
\includegraphics[width=\textwidth]{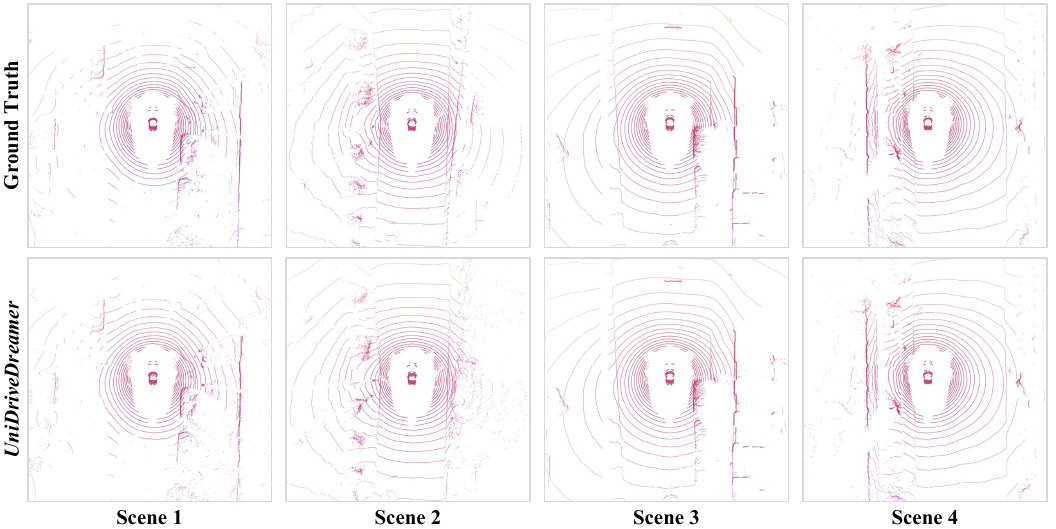}
\caption{Qualitative visualization of LiDAR reconstruction. The upper row shows the real LiDAR data, while the lower row presents the reconstructed point clouds from the LiDAR‑specific VAE in \textit{UniDriveDreamer}.}
\label{fig:lidar_recon}
\end{figure}

\noindent \textbf{LiDAR Reconstruction Results.} 
As shown in Tab. \ref{tab:recon_cmp}, we compare our LiDAR VAE against OmniGen \citep{omnigen} on LiDAR sequence reconstruction. The results demonstrate that our LiDAR VAE substantially outperforms the current SOTA in reconstruction fidelity, achieving a Chamfer Distance of 0.154 and an F‑Score of 0.900. This corresponds to an improvement of approximately 80.6\% in Chamfer Distance and 21.3\% in F‑Score over the baseline. Fig. \ref{fig:lidar_recon} provides a visual comparison  across several distinct driving scenarios between our reconstructed LiDAR sweeps and the ground truth. The visualization clearly demonstrates that the proposed LiDAR VAE preserves geometric structures effectively, producing dense and coherent point clouds that closely match the original scans.

\begin{table}[!h]
    \caption{Comparison of the reconstruction quality on nuScenes validation set.}
    \centering
    \begin{tabular}{@{}m{5cm}
        >{\centering\arraybackslash}p{3cm}
        >{\centering\arraybackslash}p{2cm}@{}}
    \toprule
    Method & Chamfer Distance $\downarrow$ & F-Score $\uparrow$ \\ 
    \midrule
    OmniGen \citep{omnigen} &0.793  &0.742\\
    \textit{UniDriveDreamer} &\textbf{0.154} &\textbf{0.900}\\
    \bottomrule
    \end{tabular}
    \label{tab:recon_cmp}
\end{table}

\noindent \textbf{Downstream Task Results.}
Since the generated data lack LiDAR intensity information, we set the intensity channel to zero for all data points during downstream task training. Using this preprocessed real data, we train BEVFusion \citep{bevfusion} from scratch to establish our baseline. As shown in Tab. \ref{tab:downstream}, the data generated by \textit{UniDriveDreamer} leads to notable improvements in downstream perception performance, increasing mAP by +0.8 and NDS by +0.52 compared to training with real data alone. These results validate the effectiveness of \textit{UniDriveDreamer} in enhancing perception performance for autonomous driving.

\begin{table}[h]
    \caption{Multimodal Generation Data Augmentation for 3D Object Detection.}
    \centering
    \begin{tabular}{@{}m{6cm}
        >{\centering\arraybackslash}p{2cm}
        >{\centering\arraybackslash}p{2cm}@{}}
    \toprule
    Training Data & mAP $\uparrow$ & NDS $\uparrow$ \\ 
    \midrule
    Original Data  &66.38  &70.01\\
    Augmented Data by \textit{UniDriveDreamer} &\textbf{67.18} &\textbf{70.53}\\
    \bottomrule
    \end{tabular}
    \label{tab:downstream}
\end{table}

\subsection{Ablation Study}
In this section, we conduct ablation studies to analyze two key design choices: (1) the effect of range‑map height repetition on LiDAR reconstruction fidelity, and (2) the impact of Unified Latent Anchoring (ULA) on cross‑modal alignment and generation quality.

\begin{figure}[!t]
\centering
\captionsetup{type=figure, justification=justified, singlelinecheck=false}
\includegraphics[width=\textwidth]{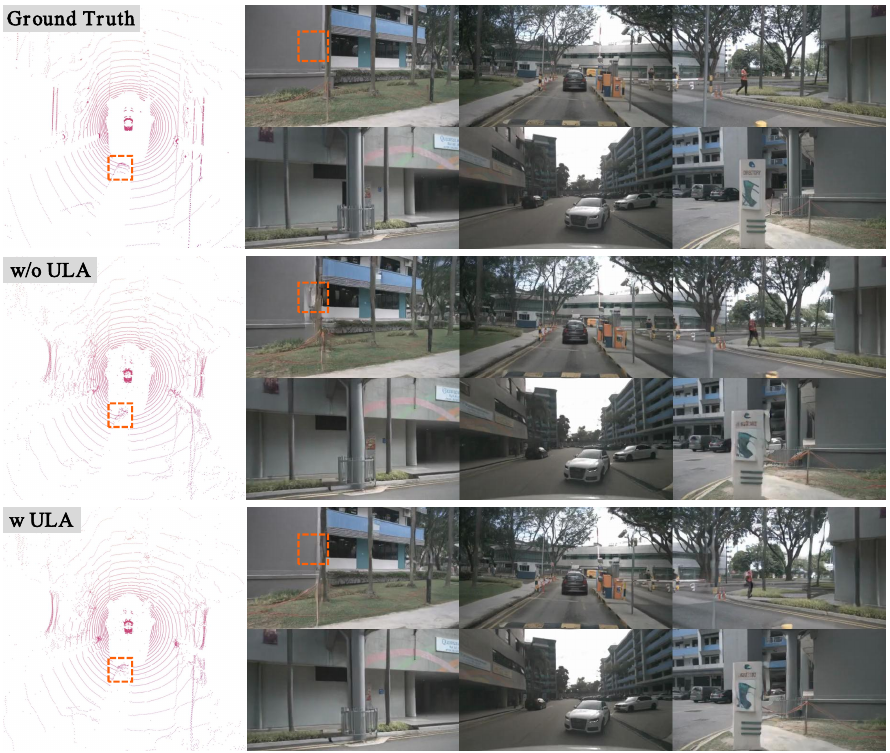}
\caption{Qualitative comparison of generated outputs with and without the Unified Latent Anchoring (ULA) module. The top row displays the ground truth. The middle row illustrates results without ULA, highlighting geometric inconsistencies and cross-modal misalignment between LiDAR points and RGB content. The bottom row presents results with ULA,showing enhanced cross‑modal geometric coherence and improved visual quality.}
\label{fig:ULA}
\end{figure}

\noindent \textbf{Effect of Range-Map Repeat Height.} We investigate the influence of height repetition in range‑map encoding on LiDAR reconstruction quality. As reported in Tab. \ref{tab:ablation_lidarvae}, progressively increasing the number of height repetitions from 1 to 4 leads to consistent improvements in Chamfer Distance and F‑Score. The best performance is achieved with 4 repetitions, reaching a Chamfer Distance of 0.154 and an F‑Score of 0.900, which validates that repeated height channels effectively mitigate range‑map sparsity and improve geometric detail recovery. Therefore, we adopt 4 repetitions as the default setting in our model.

\begin{table}[!h]
    \caption{Effect of range-map repeat height for LiDAR reconstruction fidelity.}
    \centering
    \begin{tabular}{@{}>{\centering\arraybackslash}p{3cm}
        >{\centering\arraybackslash}p{3cm}
        >{\centering\arraybackslash}p{2cm}@{}}
    \toprule
    Repeat Height & Chamfer Distance $\downarrow$ & F-Score $\uparrow$ \\ 
    \midrule
    1  &0.513 &0.794\\
    2  &0.336 &0.829 \\
    4 &\textbf{0.154} &\textbf{0.900}\\
    \bottomrule
    \end{tabular}
    \label{tab:ablation_lidarvae}
\end{table}
\noindent \textbf{Effect of ULA.} We ablate the proposed Unified Latent Anchoring (ULA) module to examine its contribution to training stability and cross-modal alignment. As illustrated in Fig. \ref{fig:ULA}, removing ULA leads to noticeable degraded consistency between the RGB and LiDAR outputs, demonstrating that explicit latent alignment enhances both intra-modal quality and cross-modal coherence. These results validate the importance of ULA in enabling robust joint multimodal learning.

\section{Conclusion}
This paper introduces \textit{UniDriveDreamer}, a single-stage unified multimodal world model. By adopting range images as the LiDAR representation, designing a LiDAR-specific VAE, and proposing Unified Latent Anchoring (ULA) for explicit cross-modal distribution alignment, our model effectively bridges the modality gap between camera and LiDAR data. Built upon a shared diffusion transformer, the model directly synthesizes temporally consistent and geometrically coherent multi‑camera videos and LiDAR sweeps, eliminating the need for cascaded or intermediate representations. Extensive experiments validate that \textit{UniDriveDreamer} surpasses SOTA methods in both camera and LiDAR synthesis. It achieves an FID of 2.81 and an FVD of 11.44 for video generation, and attains an MMD of 0.27 and a JSD of 0.039 for LiDAR generation, corresponding to relative improvements of 82.3\% in MMD and 45.8\% in JSD over the best prior work. Moreover, downstream perception tasks benefit measurably from the synthesized multimodal data, with improvements of +0.8 mAP and +0.52 NDS, confirming its practical utility for enhancing autonomous driving systems. Our work provides an efficient paradigm for multimodal sensor simulation, paving the way for more data-efficient and robust autonomous driving development.

\clearpage
\setcitestyle{numbers}
\bibliographystyle{plainnat}
\bibliography{main}

\end{document}